\documentclass[letterpaper, 10 pt, conference]{ieeeconf}

\IEEEoverridecommandlockouts                              

\usepackage{booktabs} 
\usepackage{microtype}
\usepackage{xspace}

\usepackage{array}
\usepackage{dsfont}
\usepackage{pifont}
\usepackage{times}
\usepackage{helvet}
\usepackage{courier}
\usepackage{color}
\usepackage{colortbl}
\usepackage{mathrsfs}
\usepackage{amscd}
\usepackage{stmaryrd}
\usepackage{amssymb}
\usepackage{amsmath}
\usepackage{graphicx}
\usepackage{tikz}
\usepackage{algorithm} 
\usepackage[noend]{algpseudocode}
\algrenewcommand\algorithmicindent{0.750em}%
\usepackage{subfig}
\usepackage{multirow}
\newcommand{\networks}{\mathcal{N}}

\newcommand{\distance}[2]{ ||#1||_{#2}}

\newcommand{\videoframe}{\alpha}

\newcommand{\real}{\mathbb{R}}

\newcommand{\trackset}{\mathcal{R}}
\newcommand{\track}{{\vec r}}

\newcommand{\location}{l}
\newcommand{\diff}{dist}
\newcommand{\adv}{adv}
\newcommand{\locationDistance}{\diff_{loc}}
\newcommand{\trackDistance}{\diff_{tr}}

\newcommand{\klstate}{s}
\newcommand{\klcov}{\boldsymbol{C}}
\newcommand{\estimatedlocation}{\hat{l}}
\newcommand{\attacked}[1]{#1'}

\newcommand{\commentout}[1]{}

\title{Reliability Validation of 
Learning Enabled Vehicle Tracking\thanks{Youcheng Sun and Yifan Zhou contribute equally to this work.}}

\author{Youcheng Sun$^{1*}$ and Yifan Zhou$^{2*}$, Simon Maskell$^2$ and James Sharp$^3$ and Xiaowei Huang$^2$
\thanks{$^{1}$
School of Electrical Engineering, Electronics and Computer Science, Queen's University Belfast, UK}
\thanks{$^{2}$
School of Electrical Engineering, Electronics and Computer Science, University of Liverpool, UK}
\thanks{$^{3}$
Defence Science and Technology Laboratory (Dstl), UK 
}
}

\begin{document}

\maketitle
\thispagestyle{empty}
\pagestyle{empty}

\begin{abstract}

This paper studies the reliability of a 
real-world learning-enabled system, which conducts dynamic vehicle tracking based on a high-resolution wide-area motion imagery input. The system consists of   multiple neural network components -- to process the imagery inputs -- and multiple symbolic (Kalman filter) components -- to analyse the processed information for vehicle tracking. It is known that neural networks suffer from
adversarial examples, which make them lack robustness. However, it is unclear if and how the adversarial examples over learning components can affect the overall system-level
reliability. 
By integrating a coverage-guided neural network testing tool, DeepConcolic, with the vehicle tracking system, we found that (1) the overall system can be resilient to some adversarial examples thanks to the existence of other components, and (2) the overall system presents an extra level of uncertainty which cannot be determined by analysing the deep learning components only. This research suggests the need for novel verification and validation methods for learning-enabled systems. 
\end{abstract}


\section{Introduction}
\label{sec:introduction}

The wide-scale deployment of Deep Neural Networks (DNNs) in 
safety-critical applications, such as self-driving cars, healthcare and Unmanned Air Vehicles (UAVs),  
increases the demand for 
tools that can test, validate, verify, and ultimately certify such systems~\cite{huang2018safety}. Normally, autonomous systems, or more specifically learning-enabled systems, contain both data-driven learning components and model-based non-learning components \cite{Sifakis2017}, and a certification needs to consider both categories of components, and the system as a whole.   

Structural code coverage combined with requirements coverage has been the primary approach for measuring test completeness as 
assurance evidence to support safety arguments, for the certification of
safety-critical software. Testing techniques for machine learning components, primarily DNNs, are comparatively new and have only been actively developed in the past few years, e.g., \cite{PCYJ2017,sun2018testing}. Unlike model-based software systems, DNNs are usually considered as black boxes, and therefore it is difficult to
understand their behaviour by means of inspection. 
In \cite{sun-concolic}, we developed a tool, DeepConcolic, to work with a number of extensions to the MC/DC coverage metric, targeting DNNs. The MC/DC coverage metric \cite{HVCR2001} is recommended in a number of certification documents and standards, such as RTCA's DO-178C and ISO26262, for the development of safety critical software. Its extension to DNN testing has been shown to be successful in testing DNNs by utilising white-box testing methods, i.e., by exercising the known structure with  the parameters of a DNN to 
gather assurance evidence. 
It is, however, unclear whether such a testing tool can still be effective when working with learning-enabled systems containing both learning and model-based components. Primarily, we want to understand the following two research questions: 

\begin{itemize}
    \item[Q1] Can the system as a whole be resilient against the deficits discovered over the learning components? 
    \item[Q2] Is there new uncertainty needed to be considered in terms of the interaction between learning and non-learning components? 
\end{itemize}

The key motivation for considering Q1 is to understand whether, when generating a test suite, a DNN testing tool needs to consider the existence of other components in order to assess the safety of the system.  The key motivation for considering Q2 is to understand whether a DNN testing tool can take advantage of the uncertainty presented in the interaction between learning and non-learning components in generating test cases. 


Specifically, for the learning-enabled systems, we consider in this paper a tracking system in Wide Area Motion Imagery (WAMI)~\cite{ZM2019}, where the vehicle detection is implemented by a few DNNs and the tracking is implemented with a Kalman filter based method. Using this system, we consider its reliability when running in an adversarial environment, where an adversary can have limited access to the system by intercepting the inputs to the perception unit. 

Our experiments provide affirmative answers to the above research questions and point out the urgent need to develop system-level testing tools to support the certification of learning-enabled systems.   


\section{Preliminaries}\label{sec:preliminary}

A (feedforward and deep) neural network $\networks$ is a function that maps an input $x$ to an output. According to the tasks, the output can be of different format. For example, for classification task, the DNN computes a \emph{label}, which is denoted by $\networks(x).\mathit{label}$.

\emph{Adversarial examples}~\cite{SZSBEGF2014} are two 
very similar inputs with different labels. The existence of such pairs has been used
as a proxy metric for the training quality of a DNN. 
Given an input $x_1$ that is correctly labeled by a DNN $\networks$, another input $x_2$
is said to be an adversarial example if $x_1$ and $x_2$ are ``close enough'', i.e.,
$||x_1-x_2||_p\leq \epsilon$, and $\networks(x_1).\mathit{label}\neq \networks(x_2).\mathit{label}$. Here,
$p$ denotes the $L^p$-norm distance metric and $\epsilon$ measures the term ``sufficiently small''.

A number of algorithms
have been proposed to find adversarial examples. However, such methods are not able to quantify progress, and thus, are not useful as a stopping condition for testing a DNN. 
This has motivated the coverage criteria recently developed for DNNs.

\subsection{The DeepConcolic Tool}

Several structural coverage criteria have been designed for DNNs, including neuron coverage~\cite{PCYJ2017} and its extensions~\cite{ma2018deepgauge}, and variants of Multiple Condition/Decision Coverage (MC/DC) for DNNs~\cite{sun2018testing-b}. These coverage criteria quantify the exhaustiveness of a test suite for a DNN. Neuron coverage requires that, for every neuron
in the network, there must exist at least one test case in the generated test suite that lifts its activation value above some threshold; the criteria in~\cite{ma2018deepgauge} generalise the neuron coverage from a single neuron to a set of neurons. The MC/DC variants
for DNNs capture the fact that a (decision) feature (a set of neurons)
in a layer is directly decided by its connected (condition) features 
in the previous layer, and 
thus require
test conditions such that every condition feature must be exhibited 
regardless
of its effect on the decision feature.

DeepConcolic\footnote{{https://github.com/TrustAI/DeepConcolic}} implements a concolic analysis that 
examines the set of behaviors of a DNN, and is able to identify potentially problematic input/output pairs. 
DeepConcolic
generates test cases and adversarial 
examples for DNNs following the
specified test conditions. Developers can use the test results to compare different DNN models, and the
adversarial examples can be used to improve and re-train the DNN or
develop an adversarial example mitigation strategy. Moreover, a major safety challenge for the use of DNNs
is due to the lack of understanding about how a decision is made by a DNN. The DeepConcolic tool is able to
test each particular component of a DNN, and this 
aids human analysis of
the internal structures of a DNN. By understanding these structures, we improve the
confidence in DNN behaviour; this is different from providing guarantees (for example \cite{HKWW2017,WHK2018,RHK2018,RWSHKK2019,WU2020298}), just as testing
conventional software does not provide guarantees of correctness.

\section{Detection from Wide-Area Motion Imagery}
\label{sec:wami-detector}

This section describes the technical details of a tracking system.
%
%
The tracking system requires continuous imagery input from e.g., airborne high-resolution cameras. The input to the tracking system is a video, which consists of a finite sequence of images. Each image contains a number of vehicles. 
Similar to \cite{ZM2019},  we use the WPAFB 2009~\cite{wpafbdataset} dataset. The images were taken by a camera system with six optical sensors that had already been stitched to cover a wide area of around $35\, km^{2}$. The frame rate is 1.25Hz. This dataset includes 1025 frames, which is around 13 minutes of video. It is divided into training video ($512$ frames) and testing video ($513$ frames). All the vehicles and their trajectories are manually annotated. There are multiple resolutions of videos in the dataset. For the experiment, we chose to use the $12$$,$$000$$\times$$10$$,$$000$ images, in which the size of vehicles is smaller than $10$$\times$$10$ pixels. We use $\videoframe_i$ to denote the $i$-th frame and $\videoframe_i(x,y)$ the  pixel on the intersection of $x$-th column and $y$-th row of  $\videoframe_i$. 


In the following, we explain how the tracking system works by having a video as input. In general, this is done in two steps: detection and tracking. In Section~\ref{sec:registration} through to Section~\ref{sec:framework}, we explain the detection steps, i.e., how to detect a vehicle with CNN-based perception units. This is followed by the tracking step in Section~\ref{sec:wami-tracker}. 

\subsection{Background Construction}\label{sec:registration}

Vehicle detection in WAMI video is a challenging task due to the lack of vehicle appearances and the existence of frequent pixel noises. It has been discussed in \cite{SommerTSB16,LaLondeZS18} that an
\emph{appearance-based} object detector may cause a large number of false alarms. For this reason, in this paper, we only consider \emph{detecting moving objects} for tracking. 


Background construction is a fundamental step in extracting pixel changes from the input image.
The background is built for the current environment from a number of previous frames that were captured by a moving camera system. It proceeds in the following steps. 

\paragraph{Image registration} is  to compensate for the camera motion by aligning all the previous frames to the current frame. The key is to estimate a transformation matrix, $h_{t}^{t-k}$, which transforms frame $\videoframe_{t-k}$ to frame $\videoframe_t$ using a given transformation function. For the transformation function, we consider projective transformation (or homography), which has been widely applied in multi-perspective geometry, an area where WAMI camera systems are already utilised.

The estimation of 
$h_{t}^{t-k}$ is generated by applying  
feature-based approaches.
First of all, feature points from 
images at frame $\videoframe_{t-k}$ and $\videoframe_t$, respectively, are extracted by feature detectors (e.g., Harris corner or SIFT-like~\cite{SIFT} approaches). Second, feature descriptors, such as SURF~\cite{bay2006surf} and ORB~\cite{rublee2011orb} are computed for all detected feature points. Finally, pairs of corresponding feature points between two images can be identified and the 
matrix $h_{t}^{t-k}$ can be estimated by using RANSAC~\cite{fischler1981random} which is robust against outliers.


\paragraph{Background Modeling} We generate the background, $I_{t}^{bg}$, for each time $t$, by computing the median image of the $L$ previously-aligned frames, i.e., 
\begin{equation}
I_{t}^{bg}(x,y) = \frac{1}{L}\sum_{i=1}^L \videoframe_{t-i}(x,y)
\end{equation}
In our experiments, we take either  $L=4$ or $L=5$.

Note that, to align the $L$ previous frames to the newly received frame, only one image registration process is performed. After obtaining the 
matrices $h_{t-1}^{t-2}, h_{t-1}^{t-3}, ...$ by processing previous frames, we perform image registration once to get $h_{t}^{t-1}$, and then let 
\begin{equation}\label{equ:updatetemplate}
h_{t}^{t-2}=h_{t}^{t-1} \times h_{t-1}^{t-2},~  h_{t}^{t-3}=h_{t}^{t-1} \times h_{t-1}^{t-3}.
\end{equation}

\paragraph{Extraction of Potential Moving Objects} By comparing the difference between $I_t^{bg}$ and the current frame $\alpha_t$, we can extract a set $Q_{bc}$ of potential moving objects  by first computing the following set of pixels
\begin{equation}
    P_{bc} = \{ (x,y)~|~ |I_t^{bg}(x,y) - \alpha_t(x,y)| > \delta_{bc}, (x,y) \in \Gamma \}
\end{equation}
and then applying image morphology operation on
$P_{bc}$, where $\Gamma$ is the set of pixels and  $\delta_{bc}$ is a threshold value to determine which pixels should be considered. 


\subsection{CNN for Detection Refinement}\label{sec:refinement}

After obtaining $P_{bc}$, we develop a Convolutional Neural Network (CNN) $\networks_{dr}$, a type of DNN, to detect vehicles. 
We highlight a few design decisions. 
The major causes of false alarms generated by the background subtraction are: poor image registration, light changes and the apparent displacement of high objects (e.g., buildings and trees) caused by parallax. We emphasise that the objects of interest (e.g., vehicles) mostly, but not exclusively, appear on roads. Moreover, we perceive that a moving object generates a temporal pattern (e.g., a track) that can be exploited to discern whether or not a detection is an object of interest. Thus, in addition to the shape of the vehicle in the current frame, we assert that the historical context of the same place can help to distinguish the objects of interest and false alarms. 

By the above observations, we create a binary classification CNN $\networks_{dr}:\real^{21\times 21\times (N+1)}\xrightarrow{} \{0,1\}$ to predict whether a $21 \times 21$ pixels window contains a moving object given aligned image patches generated from 
the previous $N$ frames. The $21 \times 21$ pixels window is identified by considering the image patches from the set $Q_{bc}$. We suggest $N=3$ in this paper, as it is the maximum time for a vehicle to cross the window. The input to the CNN is a $21 \times 21 \times (N+1)$ matrix and the convolutional layers are identical to the traditional 2D CNNs except that the three colour channels are substituted with $N+1$ grey-level frames. 

Essentially, $\networks_{dr}$ acts as a filter to remove from $Q_{bc}$ objects that are unlikely to be vehicles. Let $Q_{dr}$ be the obtained set of moving objects. If the size of an image patch in $Q_{dr}$ is similar to a vehicle, we directly label it as a vehicle. On the other hand, if the size of the image patch in $Q_{dr}$ is larger than a vehicle, i.e., there may be multiple vehicles, we pass this image patch to the location prediction for further processing.

\subsection{CNN for Location Prediction}\label{sec:position}

We take a regression CNN  $\networks_{lp}:\real^{45\times 45\times (N+1)}\xrightarrow{}\real^{15\times 15}$ to process image patches passed over from the detection refinement phase.
As in~\cite{LaLondeZS18}, a regression CNN can predict the locations of objects given spatial and temporal information.  
The input to $\networks_{lp}$ is similar to the classification CNN $\networks_{dr}$ described in Section~\ref{sec:refinement}, except that the size of the window is enlarged  to $45 \times 45$. 
The output of $\networks_{lp}$ is a $225$-dimensional vector, equivalent to a down-sampled image ($15 \times 15$) for reducing computational cost. 

For each $15 \times 15$ image, we apply a filter to obtain those pixels whose values are greater than not only a threshold value $\delta_{lp}$ but also the values of its adjacent pixels. We then obtain another $15 \times 15$ image with a few bright pixels, each of which is labelled as a vehicle. Let $O$ be the set of moving objects updated from $Q_{dr}$ after applying location prediction.

\subsection{Detection Framework}\label{sec:framework}

\begin{figure*}[h!]
\centering
\includegraphics[width=\textwidth]{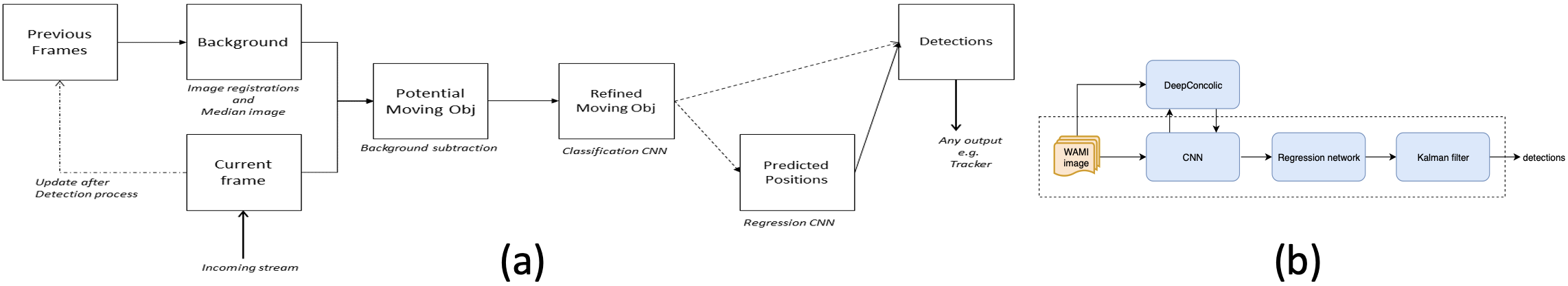}
\caption{(a) The architecture of the vehicle detector.
(b) Workflow for testing the WAMI tracking system. \label{fig:detector}}
\end{figure*}
The processing chain of the detector is shown in Figure~\ref{fig:detector}(a). At the beginning of the video, the detector takes the first $L$ frames to construct the background, thus the detections from frame $L+1$ can be generated. After the detection process finishes in each iteration, it is added to the template of previous frames. The updating process substitutes the oldest frame with the input frame. This is to ensure that the background always considers the latest scene, since the frame rate is usually low in WAMI videos such that parallax effects and light changes can be pronounced. As we wish to detect very small and unclear vehicles, we apply a small background subtraction threshold and a minimum blob size. This, therefore, leads to a huge number of potential blobs. The classification CNN is used to accept a number of blobs. As mentioned in Section~\ref{sec:refinement}, the CNN only predicts if the window contains a moving object or not. According to our experiments, the cases where multiple blobs belong to one vehicle and one blob includes multiple vehicles, occur frequently. Thus, we design two corresponding scenarios: the blob is very close to another blob(s); the size of the blob is larger than $20 \times 20$. If any blob follows either of the two scenarios, we do not consider the blob for output. The regression CNN (Section~\ref{sec:position}) is performed on these scenarios to predict the locations of the vehicles in the corresponding region, and a default blob will be given. If the blob does not follow any of the 
scenarios, this blob will be outputted directly as a detection. Finally, the detected vehicles include the output of both sets.

\subsection{Object Tracking}
\label{sec:wami-tracker}

\subsubsection{Problem Statement}

We consider a single target tracker (ie. Kalman filter) to track a vehicle given all the detection points over time in the field of view. The track is initialised by manually giving a starting point and a zero initial velocity, such that the state vector is defined as $s_{t}=[x_t, y_t, 0, 0]^T$ where $[x_t, y_t]$ is the coordinate of the starting point. We define the initial covariance of the target, $P=diag \left[30, 30, 20, 20 \right]^2$, which is the initial uncertainty of the target's state\footnote{With this configuration 
it is not necessary for the starting point to be a precise position of a vehicle, and the
tracker will find a proximate target to track on. However, it is possible to define a specific velocity and reduce the uncertainty in $P$, so the tracker can track a particular target.}.

A near-constant velocity model is applied as the dynamic model in the Kalman filter which is defined in~(\ref{eq:dynamics-model}).
\begin{equation}
s_{t|t-1} = F \cdot s_{t-1} + \omega_t~~~\text{ s.t. }    F =
\left[ 
\begin{array}{cccc}
    \mathbb{I}_{2\times2} & \mathbb{I}_{2\times2} \\
    \mathbb{O}_{2\times2} & \mathbb{I}_{2\times2} \\
\end{array}
\right]
\label{eq:dynamics-model}
\end{equation}
\noindent where $\mathbb{I}$ is a identity matrix, $\mathbb{O}$ is a zero matrix, $s_{t-1}$ is the state vector in previous timestep, $s_{t|t-1}$ is the predicted state vector and $\omega_t$ is the process noise which can be further defined as $\omega_t \thicksim \mathcal{N}(0, Q)$, where $\mathcal{N}(0, Q)$ denotes a Gaussian distribution whose mean is zero and the covariance is Q defined in~(\ref{eq:processnoiseQ}).
\begin{equation}
Q = \sigma_{q}^{2} \cdot 
    \begin{bmatrix}
    \frac{1}{3} dt^{3} \cdot \mathbb{I}_{2\times2} & \frac{1}{2} dt^{2}  \cdot \mathbb{I}_{2\times2} \\
    \\
    \frac{1}{2} dt^{2} \cdot \mathbb{I}_{2\times2} &  \mathbb{I}_{2\times2}
    \end{bmatrix}
\label{eq:processnoiseQ}
\end{equation}
\noindent where $dt$ is the time interval between two frames and $\sigma_q$ is a configurable constant. $\sigma_q=3$ is suggested for the aforementioned WAMI video.

Next, we define the measurement model as~(\ref{eq:measurementmodel}).
\begin{equation}
z_t = H \cdot s_{t} + \epsilon_t \text{\quad s.t. \quad}H = \left[ 
\begin{array}{cccc}
    1 & 0 & 0 & 0 \\
    0 & 1 & 0 & 0 \\
\end{array}
\right]
\label{eq:measurementmodel}
\end{equation}
\noindent where $z_t$ is the measurement (which is the position of the tracked vehicle), $s_t$ denotes the true state of the vehicle and $\epsilon_t\thicksim \mathcal{N}(0, R)$ denotes the measurement noise which models the uncertainty involved in the detection process. R is defined as $R = \sigma_{r}^{2} \cdot \mathbb{I}_{2\times2}$, where we suggest $\sigma_{r}=5$ for the WAMI video.

Since the camera system is moving, the position should be compensated for such motion using the identical transformation function for image registration. However, we ignore the influence to the velocity as it is relatively small, but consider integrating this into the process noise.

\subsubsection{Measurement Association}

During the update step of the Kalman filter, the residual measurement should be calculated by subtracting the measurement ($z_t$) from the predicted state ($s_{t|t-1}$). In the tracking system, the gated nearest neighbour to obtain the measurement from a set of detections is considered. K-nearest neighbour is firstly applied to find the nearest detection, $\hat z_t$, of the predicted measurement, $H \cdot s_{t|t-1}$. Then the Mahalanobis distance between $\hat z_t$ and $H \cdot s_{t|t-1}$ is calculated as follows:
\begin{equation}
    D_{t} = \sqrt{ (\hat z_t-H \cdot s_{t|t-1})^T \cdot P_{t|t-1}^{-1} \cdot (\hat z_t-H \cdot s_{t|t-1})}
    \label{eq:Mdist}
\end{equation}
where $P_{t|t-1}$ is the Innovation covariance, which is defined within Kalman filter.

A potential measurement is adopted if $D_{t} \leq g$ with $g = 2$ in our experiment. If there is no available measurement, the update step will not be performed and the state uncertainty accumulates. It can be noticed that a large covariance leads to a large search window. Because the search window can be unreasonably large, we halt the tracking process when the trace of the covariance matrix exceeds a pre-determined value.

\section{Reliability Testing Framework}
\label{sec:testing}

We consider the reliability of the vehicle tracking system introduced above when its perception units are subject to adversarial attacks. In general, we assume that the vehicle tracking system is running in an adversarial environment and the adversary is able to intercept the inputs to the perception unit in a limited way. 

Figure~\ref{fig:detector}(b) outlines the workflow of our testing framework, where DeepConcolic is deployed for reliability testing 
of the vehicle tracking system. 
Inside the dashed block, resides the workflow of the original WAMI tracking system as described in Section \ref{sec:wami-detector}. In order to test
the reliability of this vehicle tracker, we interface it with DeepConcolic, which accepts as inputs 
the original WAMI image inputs and the convolutional network. It then generates the distortion, via MC/DC  testing, for the input image to lead the  network into failing to detect 
vehicles.


\subsection{Tracks} 

First of all, we formalise the concept of a track implemented in the WAMI tracker.
At each step (frame), every detected vehicle  can be identified with a \emph{location}, represented as a tuple,
 $\location=(x,y)$, where $x$ and $y$ are the location's horizontal and vertical
coordinates.

Given two locations $\location_1=(x_1,y_1)$ and $\location_2=(x_2,y_2)$, we assume that there is
a distance function $\locationDistance(\location_1,\location_2)$ to quantify the distance between the two
detected locations. For example, we can define the distance function as follows.
\begin{equation*}
    \locationDistance(\location_1,\location_2)=\sqrt{(x_1-x_2)^2+(y_1-y_2)^2}
\end{equation*}
In reality, when tracking a car, it is reasonable to assume a threshold $\epsilon$
such that as long as the distance between two locations does not exceed $\epsilon$, i.e., $\locationDistance(\location_1,\location_2)\leq \epsilon$, the two locations $\location_1$ and $\location_2$ can be regarded as the same.

A detected \emph{track}, denoted by $\track$, consists of a sequence
of locations such that $\track=\location_{1} \dots \location_{n}$, where
$n$ is the total number of steps for tracking.
%
We write the number of steps $|\track|$ as the length of the track. Let $\trackset$ be the set of finite tracks. 
Subsequently, given two tracks $\track_{1}$ and $\track_{2}$ of the same length, there are multiple choices to measure
the difference between the two, denoted by $\trackDistance(\track_1,\track_2)$, including
\begin{equation*}
\begin{array}{lcl}
    \displaystyle \trackDistance^{sum}(\track_1, \track_2) & = &\sum_{1\leq i\leq n}\locationDistance(\location_{1,i},\location_{2,i}) \\
    
    \displaystyle \trackDistance^{mean}(\track_1, \track_2) &=&\frac{1}{n}\cdot \diff^{sum}(\track_1, \track_2) \\
    
    \displaystyle \trackDistance^{max}(\track_1, \track_2)&=&\max_{1\leq i\leq n}\{\locationDistance(\location_{1,i},\location_{2,i})\}\\
    
    \displaystyle \trackDistance^{\epsilon}(\track_1, \track_2)&=&~|~\{\locationDistance(\location_{1,i},\location_{2,i})>\epsilon|1\leq i\leq n\}|
\end{array}
\label{eq:diff}
\end{equation*}
where $\location_{m,i}$ represents the $i$-th location on the track $\track_m$. Intuitively, $\trackDistance^{sum}(\track_1, \track_2)$ expresses the accumulated distance between locations of two tracks,  $\trackDistance^{mean}(\track_1, \track_2)$ is the average distance between locations of two tracks,   $\trackDistance^{max}(\track_1, \track_2)$ highlights the largest distance between locations of two tracks, and $\trackDistance^{\epsilon}(\track_1, \track_2)$ counts the number of steps on which the distance between locations of two tracks is greater than $\epsilon$.


\subsection{Reliability Definition}

\begin{figure}
    \centering
    \includegraphics[width=.475\columnwidth]{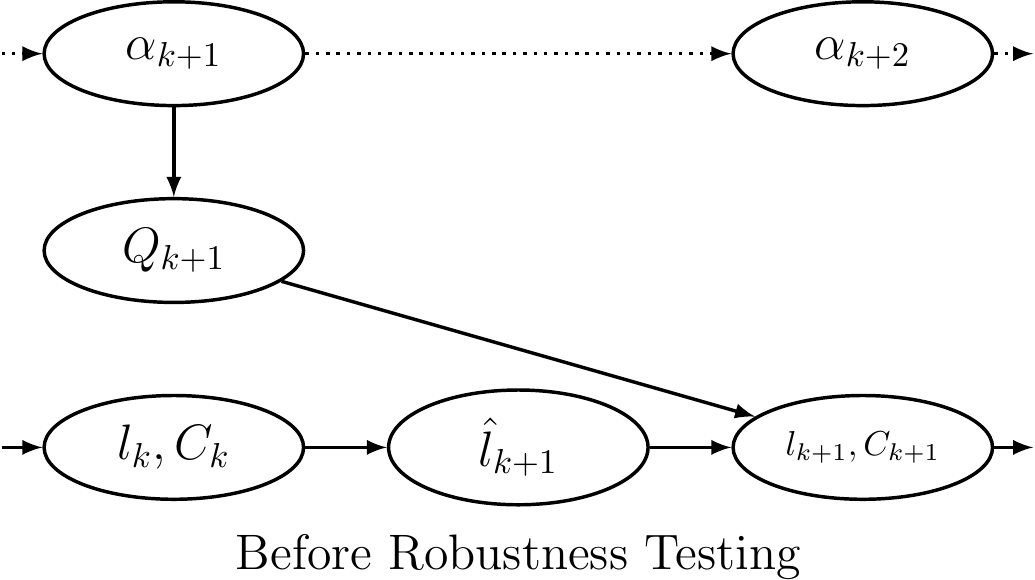}\hfill
    \includegraphics[width=.475\columnwidth]{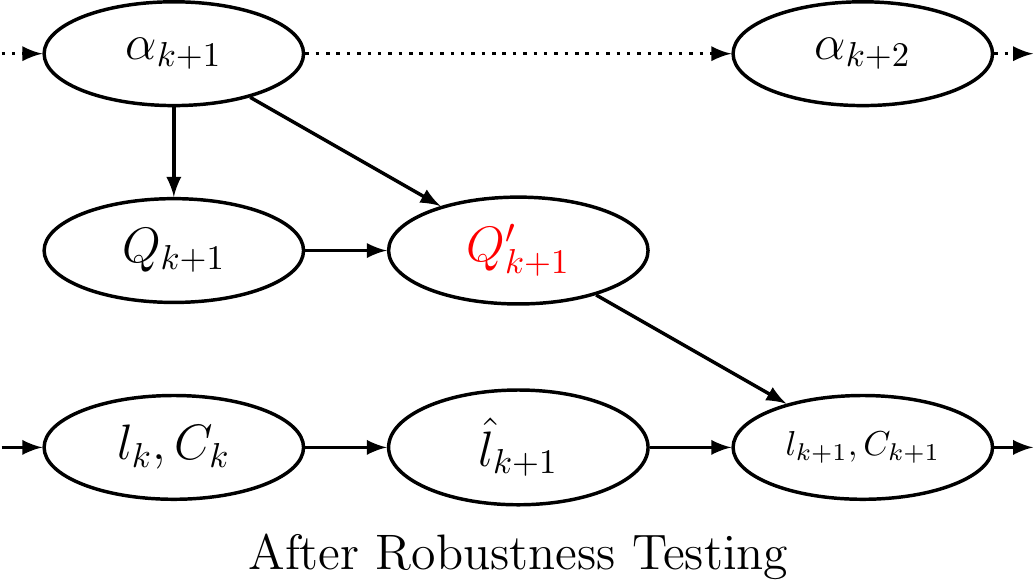}
    \caption{Illustration of the dependencies and workflows between data, for the cases of before robustness testing (Left) and after robustness testing (Right), respectively.}
    \label{fig:functions}
    \vspace*{-.5cm}
\end{figure}

We start from explaining how to obtain a track by the WAMI detector in Section~\ref{sec:wami-detector}. 
Given a number $n$, we let $X^n$ be the set of image sequences of length $n$. 
For each image sequence $\videoframe_1 \dots \videoframe_n \in X^{n}$, the WAMI tracking system maps it into a track as follows.
%
Let $\videoframe_{k+1}$ be the $(k+1)$-th image and $\klstate_k=(\location_k,\klcov_k)$ the state of the Kalman filter after $k$ steps, where $\location_k$ is the location, and $\klcov_k$ is the covariance matrix representing the uncertainty of $\location_k$. From $\klstate_k$, an estimated location $\estimatedlocation_{k+1}$ can be obtained. Moreover, from the image $\videoframe_{k+1}$, by the WAMI detection based on both $\networks_{dr}$ and $\networks_{lp}$ in Section~\ref{sec:wami-detector}, we can have a set of locations $Q_{k+1}$ representing the detected moving objects. By selecting a location $q_{k+1}\in Q_{k+1}$ which is closest to $\estimatedlocation_{k+1}$, we determine the next location $\location_{k+1}$, and use this information to update $\klcov_{k}$ into $\klcov_{k+1}$. In the left of Figure~\ref{fig:functions}, the causality between the above entities is exhibited. 
For simplicity, we can write the resulting track as 
\begin{equation}
\track = G(\videoframe_1...\videoframe_n)=\location_1 ... \location_n
\end{equation} 

Because of the existence of the adversary, when the detection components $\networks_{dr}$ and $\networks_{lp}$ lack robustness, from $\videoframe_{k+1}$ we may have a different set,  $\attacked{Q}_{k+1}$, of detected moving objects. Therefore, there may be a different $\attacked{q}_{k+1}\neq q_{k+1}$, which will in turn lead to different $\attacked{\location}_{k+1}$ and $\attacked{\klcov}_{k+1}$. The causality of these entities is exhibited in the right of Figure~\ref{fig:functions}. We write the resulting track as 
\begin{equation}
\track' = (\adv\odot G)(\alpha_1...\alpha_n)=\location_1'...\location_n'
\end{equation}

Therefore, given a  track $\track$, by attacking the learning components of the tracking system,
another track $\track'$ may be returned. 
Now, we can define the reliability of a tracking system.
The reliability of a tracking system over an image sequence, $\videoframe_1...\videoframe_n$, a function, $\trackDistance$, and a threshold, $E$, is defined as the non-existence of an adversary, $\adv$, such that
\begin{equation}
\trackDistance(G(\alpha_1...\alpha_n),(\adv\odot G)(\alpha_1...\alpha_n))>E.
\end{equation}
when $\distance{\alpha_1...\alpha_n-\adv(\alpha_1...\alpha_n)}{p} \leq \epsilon_0$, with
$\epsilon_0 > 0 $ being a small enough real number. We use
$\distance{\alpha_1...\alpha_n-\adv(\alpha_1...\alpha_n)}{p} \leq \epsilon_0$ to denote that the maximum  adversarial perturbation to be considered is no more than $\epsilon_0$, when measured with $L^p$-norm. Under such a constraint, the reliability requirement is to ensure that no adversary can produce an  attack that results in the attacked track significantly deviating from the original track, i.e., $\trackDistance(G(\alpha_1...\alpha_n),(\adv\odot G)(\alpha_1...\alpha_n))>E$.

\subsection{Reliability Validation} 


We use DeepConcolic generated test cases for the detection components $\networks_{dr}$ and $\networks_{lp}$ to validate the reliability of the tracking system. In theory, we may generate test cases for all steps $k>1$ or a few randomly selected steps. 
In practice, we found that the tracking system in Figure \ref{fig:detector} (b) is more sensitive to
the scheme when there are consecutive missing detections of the vehicle.

Thus, we design the testing scheme $Test(s,k)$ to test the tracking system, that is, starting from
the $s$-th frame, test cases are generated by applying DeepConcolic to $k$ successive frames for the tracking system. 
%
%
%
A track comprises of a finite sequence of vehicle detection, and
$Test(s,k)$ is useful to access the reliability of different parts of the track.
This helps reveal the potential vulnerabilities of the tracking system.

\section{Experiments}
\label{sec:experiments}

\begin{figure}[!htb]
  \centering
  \subfloat[]{
    \includegraphics[width=0.45\linewidth]{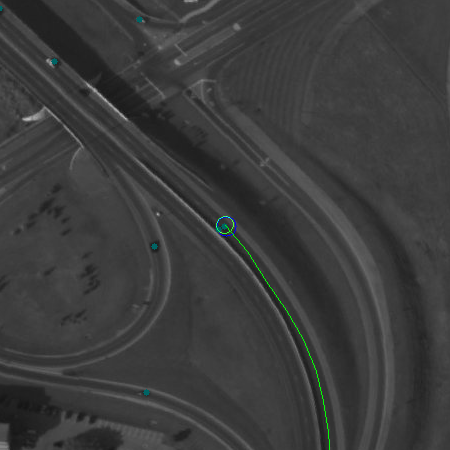}
  }\hspace{0.cm}
  \subfloat[]{
    \includegraphics[width=0.45\linewidth]{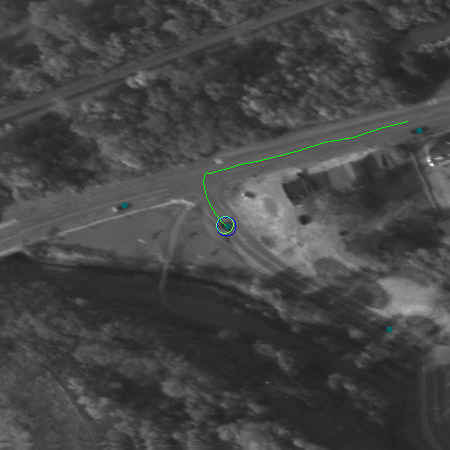}
  }\\\vspace{-0.45cm}
  \subfloat[]{
    \includegraphics[width=0.45\linewidth]{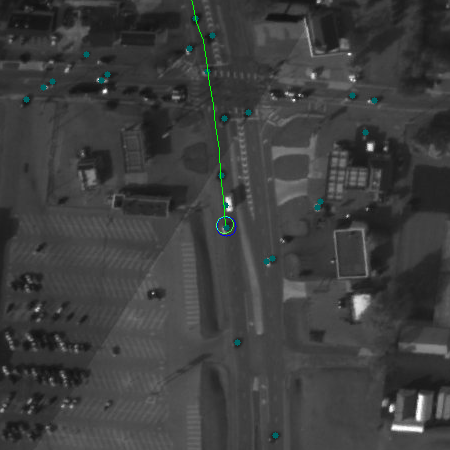}
  }\hspace{0.cm}
  \subfloat[]{
    \includegraphics[width=0.45\linewidth]{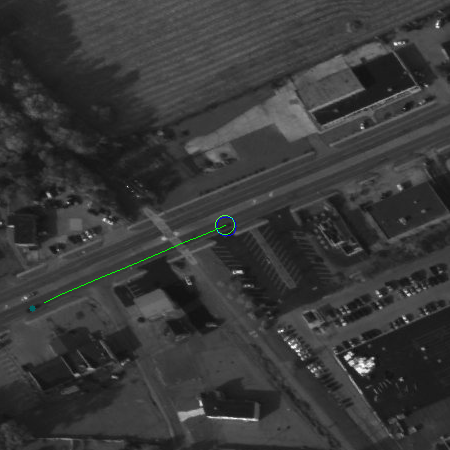}
  }
  \vspace*{-.25cm}
  \caption{Original detected tracks from the tracking system}
  \label{fig:tracks}
\end{figure}

We sample a number of tracks with maximum length of 30, and then apply the $Test(s,k)$ with a variety of configurations: $s\in[5,20]$, $k\in[1,4]$. Example  tracks are as in Figure \ref{fig:tracks}.

\begin{figure}[!htb]
  \centering
  \subfloat[]{
    \includegraphics[width=0.45\linewidth]{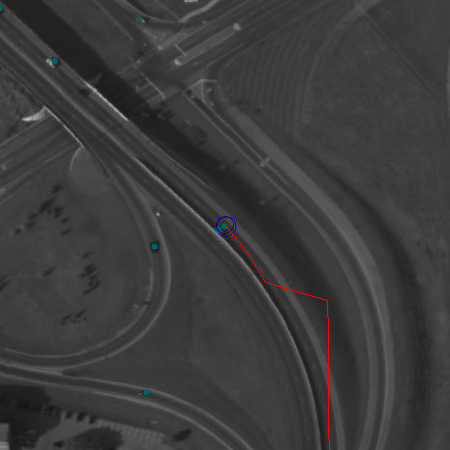}
  }\hspace{0.cm}
  \subfloat[]{
    \includegraphics[width=0.45\linewidth]{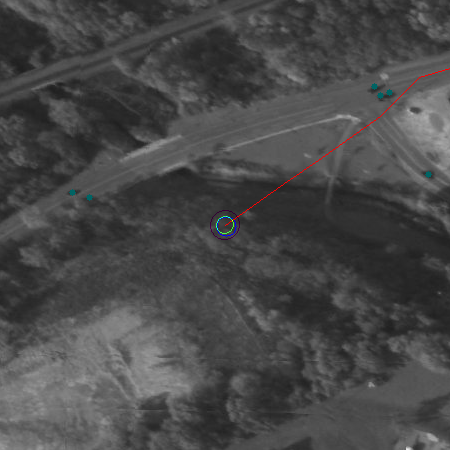}
  }\\\vspace{-0.45cm}
  \subfloat[]{
    \includegraphics[width=0.45\linewidth]{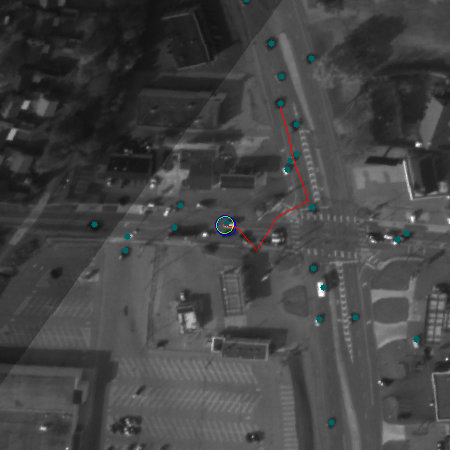}
  }\hspace{0.cm}
  \subfloat[]{
    \includegraphics[width=0.45\linewidth]{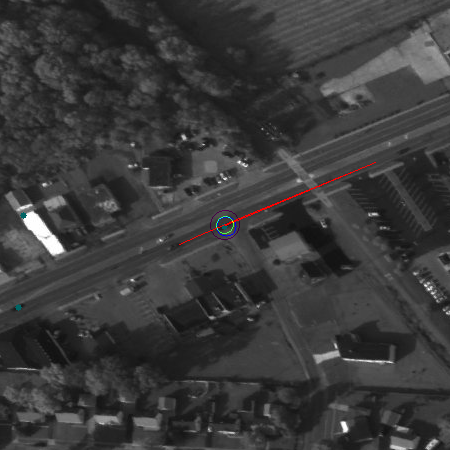}
  }    
  \vspace*{-.25cm}
  \caption{Distorted tracks found by DeepConcolic testing}
  \label{fig:adv-tracks}
    \vspace*{-.5cm}
\end{figure}

At first, we confirm that even by only testing the deep learning component
in the tracking system, it is able to find test inputs that help discover the
vulnerabilities in the tracking. Figure \ref{fig:adv-tracks} shows the adversarial
tracks found by the testing.

\begin{figure}[!htb]
  \centering
  \subfloat[$Test(5,2)$]{
    \includegraphics[width=0.45\linewidth]{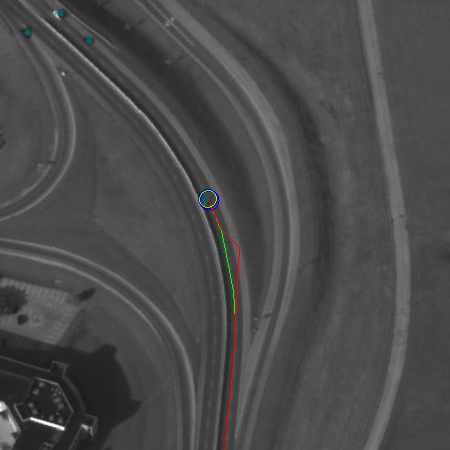}
  }\hspace{0.cm}
  \subfloat[$Test(11,2)$]{
    \includegraphics[width=0.45\linewidth]{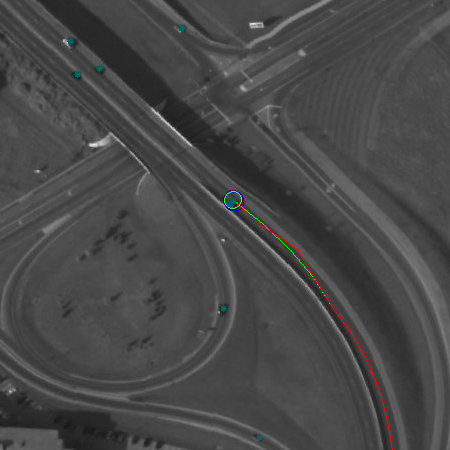}
  }\hspace{0.cm}\\
  \caption{Adversarial tracks (red) after testing different parts of the
original track (green)}
  \label{fig:q1}

\end{figure}

Furthermore, different parts of the same track
exhibit different levels of robustness. For example, as shown in Figure \ref{fig:q1},
the frames $x_5,x_6$ are less robust than $x_{11},x_{12}$, as the testing result,
from the latter, results in less deviation from the original track;
such information provides insight to evaluate and improve the tracking system.

Meanwhile, we also observe that by only testing the deep learning
component it may not be sufficient to mislead the overall tracking system. 
As in Figure \ref{fig:q1}, the tracking finally converges to the original one.
This is due to  compensation provided by other components in the system, and
it answers the research question, Q1, that the tracking system is resilient
to some extent towards the deficits of a learning component.

\begin{figure}[!htb]
  \centering
  \subfloat[$Test(9,2)$]{
    \includegraphics[width=0.45\linewidth]{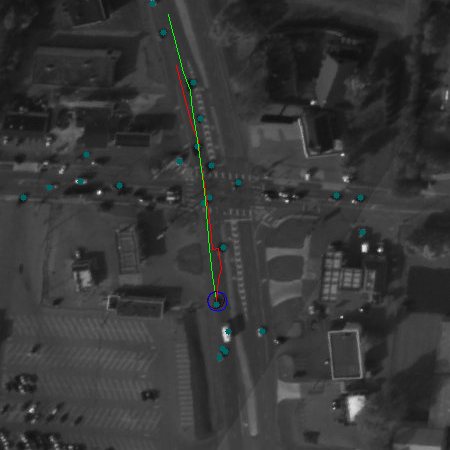}
  }
  \subfloat[$Test(9,3)$]{
    \includegraphics[width=0.45\linewidth]{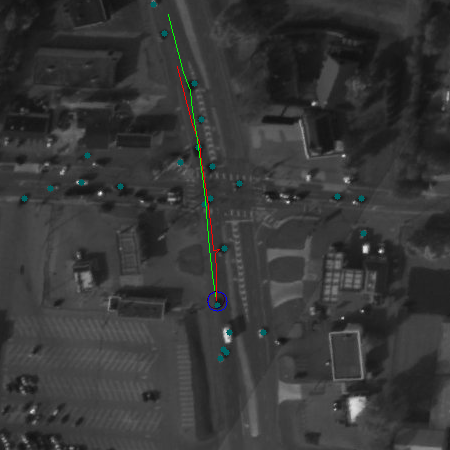}
  }
  \caption{Adversarial tracks from testing different numbers of consecutive frames starting from $x_9$}
  \label{fig:q2}
\end{figure}

Finally, changing  more frames does not necessarily result in
larger deviation from the original track. This is demonstrated as in
Figure \ref{fig:q2}, where a larger distance between the original and
adversarial tracks is found when testing frames $x_9, x_{10}$, instead of 
$x_9, x_{10}, x_{11}$. This observation reflects the uncertainty from
other components (other than the CNNs) in the tracking
system, as specified in research question Q2.

\commentout{

\subsection{Reliability Testing Results}
In this part, we are going to show how DeepConcolic can be applied to analyse the reliability of the WAMI vehicle tracking system.

TODO

\subsection{Comparison Between Two Tracking Systems}
In this part, besides the vehicle tracking system we described in the paper, we consider a second 
system such that the learning component is replaced by a (bad? not that bad?) DNN trained without using sufficient data.
Then, we compare the reliability between the two tracking systems via our testing.

}

\section{Related Works}
\label{sec:related}

Verification of DNNs from software testing perspective has been a popular direction \cite{sun-concolic,sun2019concolic,gopinath2018symbolic,odena2018tensorfuzz,xie2018coverage,wang2018adversarial,wang2018detecting,deepmutation,shenmunn,ding2017validating,dwarakanath2018identifying,deeproad}, 
see \cite{huang2018safety} for a survey with other perspectives such as formal verification and interpretability.
Autonomous driving has been the primary application domain for
assessments of DNN testing techniques~\cite{dreossi2017systematic,dreossi2017compositional,tuncali2018simulation,tuncali2018reasoning}.
The analysis in \cite{dreossi2017systematic,dreossi2017compositional} comprises an image generator
that produces synthetic pictures for testing neural networks used
in classification of cars in autonomous vehicles. In \cite{tuncali2018simulation},
a 3D simulator is proposed to test the dynamics of the pedestrians and the
agent vehicles (including simple dynamics for suspension,
tyres, etc.), in the virtual environment of the system under test.
In \cite{tuncali2018reasoning}, a technique is presented to reason about the safety of a
closed-loop, learning-enabled control system. Research is also  conducted on the verification of cognitive trust between human and autonomous systems \cite{HKO2019}.


We note that there is limited work focussed on the verification of learning-enabled systems such as the one highlighted in this work.

\section{Conclusions}
\label{sec:conclusions}

In this paper, we 
%
%
show that solely  testing the correctness of deep learning components in a vehicle tracking system in isolation is insufficient, since either a deficit discovered in the learning component may be suppressed by the existence of other components, or that there are new uncertainties introduced due to the interaction between learning and non-learning components. Similar results are also observed for the connections between LSTM and CNN layers \cite{huang2019test}. These results clearly indicate the necessity of developing a testing strategy that addresses learning components in isolation, as well as in combination with other components within a wider system-level testing framework for learning-enabled systems. 

\bigskip

{\footnotesize
This document is an overview of UK MOD (part) sponsored research and is released for informational purposes only. The contents of this document should not be interpreted as representing the views of the UK MOD, nor should it be assumed that they reflect any current or future UK MOD policy. The information contained in this document cannot supersede any statutory or contractual requirements or liabilities and is offered without prejudice or commitment.

Content includes material subject to \textcopyright~Crown copyright (2018), Dstl. This material is licensed under the terms of the Open Government Licence except where otherwise stated. To view this licence, visit {http://www.nationalarchives.gov.uk/doc/open-government-licence/version/3} or write to the Information Policy Team, The National Archives, Kew, London TW9 4DU, or email: {psi@nationalarchives.gsi.gov.uk}.

}

\bibliographystyle{unsrt}
\bibliography{all}

\end{document}